\documentclass[conference]{IEEEtran}
\IEEEoverridecommandlockouts
\usepackage{cite}
\usepackage{amsmath,amssymb,amsfonts}
\usepackage{algorithmic}
\usepackage{graphicx}
\usepackage{textcomp}
\usepackage{url}
\usepackage[tight]{subfigure}
\def\BibTeX{{\rm B\kern-.05em{\sc i\kern-.025em b}\kern-.08em
    T\kern-.1667em\lower.7ex\hbox{E}\kern-.125emX}}
\begin{document}

\title{Improving Multiple Object Tracking with Optical Flow and Edge Preprocessing}

\author{\IEEEauthorblockN{David-Alexandre Beaupr\'{e}, Guillaume-Alexandre Bilodeau}
\IEEEauthorblockA{\textit{LITIV lab., Dept. of Computer \& Software Eng.} \\
\textit{Polytechnique Montr\'{e}al}\\
Montr\'{e}al, Canada \\
\{david-alexandre.beaupre, gabilodeau\}@polymtl.ca}
\and
\IEEEauthorblockN{Nicolas Saunier}
\IEEEauthorblockA{\textit{Dept. of Civil, Geo and Mining Eng.} \\
\textit{Polytechnique Montr\'{e}al}\\
Montr\'{e}al, Canada \\
nicolas.saunier@polymtl.ca}
\thanks{This work was supported by a scholarship from the National Sciences and Engineering Research Council of Canada (NSERC) and a grant from the Fonds de recherche du Quebec - Nature and Technologies (FRQNT)}
}

\maketitle

\begin{abstract}
In this paper, we present a new method for detecting road users in an urban environment which leads to an improvement in multiple object tracking. Our method takes as an input a foreground image and improves the object detection and segmentation. This new image can be used as an input to trackers that use foreground blobs from background subtraction. The first step is to create foreground images for all the frames in an urban video. Then, starting from the original blobs of the foreground image, we merge the blobs that are close to one another and that have similar optical flow. The next step is extracting the edges of the different objects to detect multiple objects that might be very close (and be merged in the same blob) and to adjust the size of the original blobs. At the same time, we use the optical flow to detect occlusion of objects that are moving in opposite directions. Finally, we make a decision on which information we keep in order to construct a new foreground image with blobs that can be used for tracking. The system is validated on four videos of an urban traffic dataset. Our method improves the recall and precision metrics for the object detection task compared to the vanilla background subtraction method and improves the CLEAR MOT metrics in the tracking tasks for most videos.
\end{abstract}

\begin{IEEEkeywords}
object detection, object tracking, edges, optical flow, urban scenes
\end{IEEEkeywords}

\begin{figure}[t] 
	\centering
	\subfigure[] 
    {
  	\includegraphics[trim={0.75cm 0 0.75cm 0},clip, height=2.3cm]{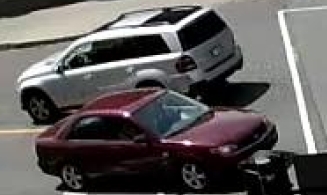}
	}
	\subfigure[] 
    {
  	\includegraphics[height=2.3cm]{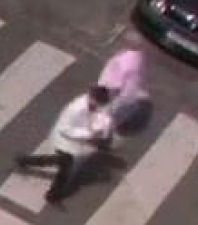} 
	}
 	\subfigure[] 
    {
  	\includegraphics[height=2.3cm]{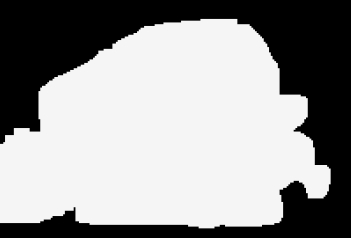} 
	}
	\subfigure[] 
    {
  	\includegraphics[trim={0.75cm 0 0.85cm 0},clip, height=2.3cm]{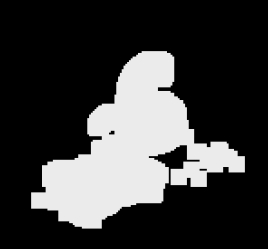} 
	}
	\subfigure[] 
    {
  	\includegraphics[trim={0cm 0 0.5cm 0},clip,height=2.3cm]{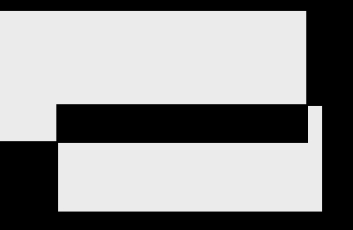} 
 	}
 	\subfigure[] 
    {
  	\includegraphics[trim={0.4cm 0 0.5cm 0},clip,height=2.3cm]{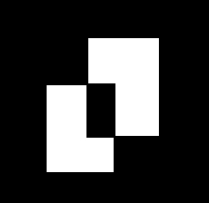}
 	}
  \caption{Example of segmentation from our method and ViBe method. a) Original image of an occlusion between cars from the Sherbrooke video sequence, b) original image of an occlusion between pedestrians from the Rouen video sequence, c) and d) respective segmentations with ViBe, and e) and f) respective segmentations from our method. Note that since the object masks that we output are combinations of bounding boxes, objects are just segmented coarsely.}
\label{fig1}
\end{figure}

\section{Introduction}
Object detection is a fundamental task in the field of computer vision. It is a necessary step in traffic surveillance in order to collect traffic data and analyze road user behavior. It is used to extract image regions that correspond to the objects of interest. Improving the detection of the cars, cyclists and pedestrians can help to improve another important task, which is multiple object tracking (MOT). Many trackers, for instance Urban Tracker (UT) \cite{DBLP:conf/wacv/JodoinBS14,7457695} and Multiple Kernelized Correlation Filter Tracker (MKCF) \cite{DBLP:journals/corr/YangB16a}, use foreground blobs $B_i$ from background subtraction as an input to track the objects in the video because these detections are generic and do not assume any prior classes. UT is a more complex tracker using feature points and a state machine to keep track of the different objects while MKCF is a fast tracker with simpler data association for tracking multiple objects. Yet, both depend on the quality of background subtraction.

There are many problems with the images produced by background subtraction methods in an urban environment. The first one is foreground blob merging, which occurs when two road users occlude each other or are close to each other as in figure \ref{fig1}. Even if trackers have ways of dealing with these occlusions, we found that it is advantageous to explicitly detect the different occluding objects prior to tracking. Another problem of background subtraction methods is the case of fragmentation where a unique object is separated in multiple smaller blobs. Once again, we found that it is easier to explicitly merge the fragmented blobs than to let the tracker decide if the multiple blobs were part of the same object or not. Another problem of background subtraction is caused by shadows, mainly by pedestrians' shadows. In fact, foreground blobs will often include the shadows of the objects, which can lead to a tracking box that is much larger and not as precise as the one without shadows, or to merging different objects in the same blob. Our method is able to eliminate most of the unwanted shadows which lead to a more precise detection. One last problem is that the background subtraction blobs are generally much bigger than the real size of the objects that we want to detect. Our proposed method was able to effectively adjust the size of the proposed blobs which results in better recall in the detection metrics.

Our method uses background subtraction \cite{DBLP:journals/tip/BarnichD11}, optical flow \cite{kroegerECCV2016} and edge processing in order to create a new binary image of foreground blobs. Background subtraction is used to locate the regions of interest (RoIs), which are the location of the foreground blobs in the current frame. The dense optical flow, with a patch size of 8, is then computed for each blob $B_i$ in the given frame. The motion vectors of the different regions and the relative distance between the regions are compared to merge the blobs that are very likely of being fragmented regions of the same object. The optical flow computed at each frame (with the previous one) is also used to separate objects merged in the same blob that are moving in opposite direction. We use the Canny edge detector \cite{Canny86acomputational} on both the blobs of the source frame and the scene background image (see section \ref{bimg}) to obtain the edges of the foreground objects (we want to eliminate background edges that might be in a foreground blob i.e. road markings near pedestrians). This last step allows adjusting the size of the objects, separating close objects that appeared as one blob in background subtraction and eliminating noise. With all this information, our novel method generates a new binary image with processing steps that handle fragmentation, merging and remove noise while giving a more precise segmentation.

The organization of the paper is the following: in section \ref{background}, we discuss related work. In section \ref{methodology}, we present our new method consisting of the foreground image, merging of similar optical flow regions, separation of opposite flow regions, edge processing and creation of the new binary image. In section \ref{results}, we present our results and finally, in section \ref{conclusion}, we conclude this paper.

\section{Related work}
\label{background}
Many methods can be used in order to extract the object RoIs in a given frame. Objects proposal methods like \cite{ZitnickECCV14edgeBoxes,BingObj2014} can get good recall results given a large number of proposals. Also, these methods do not require the input to be a video since they propose boxes based on their ``objectness''. The downside of object proposal methods is to filter the thousands of initial proposals to extract the real objects in our frame, which is often less than twenty in tracking tasks, and to make sure that every object only has one bounding box. The challenge of keeping the best box around each object while keeping high recall is difficult to achieve for the purpose of a tracking.

Another method to extract RoIs is optical flow as in \cite{kroegerECCV2016,WeinzaepfelRHS13}. Optical flow is the process of computing the motion of every pixel between two consecutive frames. By grouping pixels with similar motion, this results in blobs of pixels for each object with different motion. Thus, these methods are very good at detecting moving objects, but segmenting individual objects from a group can be more difficult, especially if they are moving in the same direction. In fact, two objects very close to one another will be considered in the same motion flow blob since their flow vectors will be very similar. In addition, these methods cannot detect still objects. However, two close objects going in opposite directions are very easy to separate with optical flow methods as stated earlier.

Recently, deep learning methods have achieved great results in object detection as seen in \cite{NIPS2015_5638,redmon2016yolo9000} while being able to make those detections almost in real time. However, these neural networks must be trained on every class we want them to detect, which can take up a lot of time and resources. They cannot detect objects from unexpected classes.

Finally, another traditional approach to obtain object RoIs is background subtraction, like with ViBe \cite{DBLP:journals/tip/BarnichD11} and SubSENSE \cite{St-Charles1107}. In this case, RoIs are the results of the differences between the current frame and a background frame model. These methods can detect objects from any class. However, they can be sensitive to camera motion and shadows. Also, they cannot resolve merging caused by occlusion or proximity. However, they are very appealing for tracking in urban scenes because of the unknown variety of objects of interest these scenes may contain. 

As mentioned above, we chose ViBe \cite{DBLP:journals/tip/BarnichD11} to provide us with the initial RoIs. ViBe is a background subtraction method that keeps track of the values of each pixels in the past to determine if a pixel in the current frame is in the foreground or the background. For a given frame, every blob $B_i$ produced by ViBe will be fed into our algorithm in order to improve the detection of objects.

\begin{figure*}
\centering
\includegraphics[width=0.95\textwidth]{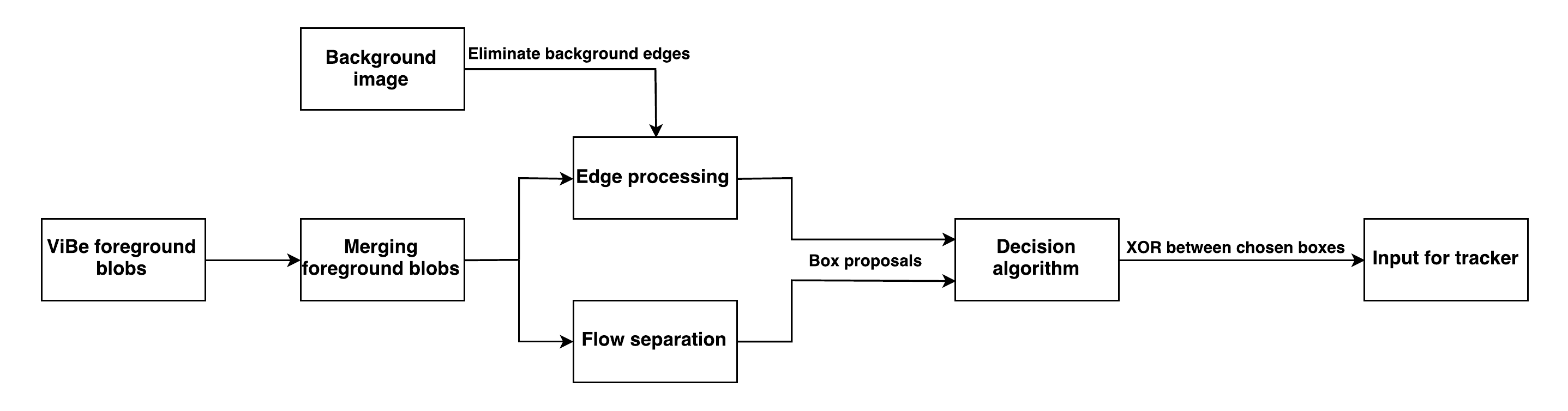}
\caption{Diagram of the steps of our method}
\label{fig2}
\end{figure*}

\section{Methodology}
\label{methodology}
This section presents the different steps of our method as shown in figure \ref{fig2}. These steps are simple operations, using optical flow and edge analysis.

\subsection{Background image}
\label{bimg}
The first operation is to accumulate a color background image from the video sequence. This will become useful in the edge processing step (see section \ref{edge proc}) because we will be able to filter out most of the background edges that may be included in foreground edges e.g. road marking that are not of interest \cite{FarahTIP2013}. The background image $A_i$ is given by
\begin{equation}
A_{i}=\alpha*I+(1-\alpha)*A_{i-1} ,
\end{equation}
where $\alpha$ is an accumulation rate and $I$ is an input image. In the experiments, $\alpha=0.01$, which means that each new frame has a weight of 0.01 in the running average and the mean image has a weight of 0.99.

\subsection{Merging foreground blobs}
\label{merging}
The first step to process a frame is to check if we can merge any foreground blobs that satisfy the three following conditions (C1), (C2) and (C3) presented below. If this is the case, a new blob $B_k$ is formed from the union of the blobs $B_i$ and $B_j$. The union operation in our case takes the smallest box that frames both blobs $B_i$ and $B_j$. 

The first condition (C1) is given by
\begin{equation}
d(B_i, B_j) \leqslant T_M ,
\end{equation}
where $d(B_i, B_j)$ is the minimum distance between the pixels of two blobs $B_i$ and $B_j$. This distance must be smaller that a threshold $T_M$ that can be modified, but we found experimentally that a distance of 7 pixels is a good compromise because we need to merge objects of various sizes (car and pedestrian dimensions vary in different datasets). 

The second condition is based on intervals of the magnitude $mag()$ of the optical flow of blobs, built as the mean magnitude $\overline{mag(B_i)}$ plus or minus one standard deviation $std(mag(B_i))$. The lower and upper bounds for the blobs $B_i$ and $B_j$ can be written as 
\begin{equation}
\begin{split}
min_{B_i} = \overline{mag(B_i)} - std(mag(B_i)) \\
max_{B_i} = \overline{mag(B_i)} + std(mag(B_i))
\end{split}
\end{equation}
\begin{equation}
\begin{split}
min_{B_j} = \overline{mag(B_j)} - std(mag(B_j)) \\
max_{B_j} = \overline{mag(B_j)} + std(mag(B_j)) .
\end{split}
\end{equation}
We can now define the domain of possible values for each blob as
\begin{equation}
\begin{split}
dom_{B_i} = \left[min_{B_i}, max_{B_i} \right] \\
dom_{B_j} = \left[min_{B_j}, max_{B_j} \right] .
\end{split}
\end{equation}
The second condition (C2) is then given by
\begin{equation}
dom_{B_i} \cap dom_{B_j} \neq \emptyset .
\end{equation}
This condition verifies that both domains $dom_{B_i}$ and $dom_{B_j}$ have at least one value in common. The third condition (C3) is
\begin{equation}
\lvert{ang(B_i) - ang(B_j)}\rvert \leqslant A_T .
\label{cond3}
\end{equation}
This condition checks if the angles $ang()$ of the optical flow for the blobs $B_i$ and $B_j$ are approximatively in the same direction. We take the angle of the optical flow at the center point of each blob for the comparison with the threshold $A_T$. The value of $A_T$ is $\frac{\pi}{2}$. This means that we sometimes merge (i.e. union operation) two foreground blobs that should not have been merged, but we prefer to err on the side of over-merging because it is possible to separate objects at a later stage of our method. At this step, we also save which RoIs were modified and store them in a map that will be used in section \ref{decision}. These new foreground blobs will be the new RoIs for the next steps.

\subsection{Flow separation}
\label{flow sep}
The purpose of this step is to separate foreground blobs that contain two objects going in opposite directions. For each foreground blob $B_i$ in the image, we apply the k-means clustering algorithm to the optical flow vectors. We chose $k=3$ because when there are two objects moving in opposite direction, the segmentation of these objects results in one cluster for the background, and the other two clusters as the distinct objects. We fit bounding boxes $r_1$, $r_2$ and $r_3$ around each of the three clusters. We can then compute the ratio, $ratio_{int}(i,j)$, of the intersection of the boxes over the smallest area $min_{area}$ of the two boxes as
\begin{equation}
ratio_{int}(i, j) = \frac{r_i \cap r_j}{min_{area}(r_i, r_j)} \text{ for every } i \neq j.
\end{equation}
Since there are only three boxes, this gives us three $ratio_{int}(i, j)$ for all pairs $(r_i, r_j)$. We compare each $ratio_{int}(i, j)$ against a threshold $T_{int}$ of 0.40 and if the ratio is smaller, we check if the boxes $r_i$ and $r_j$ are going in opposite directions using the negation of the third condition expressed by equation \ref{cond3}. For a given blob $B_k$, we will keep two bounding boxes ($r_i$ and $r_j$) if both conditions were met. If not, we simply fit a bounding box around the original foreground blob $B_k$ for this particular region and ignore $r_1$, $r_2$ and $r_3$. During this process, we save every RoI that has been split in two in a map that will be used in section \ref{decision}.

\begin{figure*}[t] 
  \centering
  \subfigure[] 
   {
  \includegraphics[height=2cm]{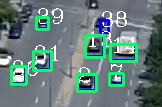} 
  }
  \subfigure[] 
   {
  \includegraphics[height=2cm]{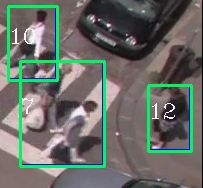}
  }
   \subfigure[] 
   {
  \includegraphics[height=2cm]{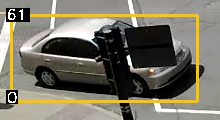} 
  }
 \subfigure[] 
   {
  \includegraphics[trim={0.15cm 0 0.15cm 0},clip,height=2cm]{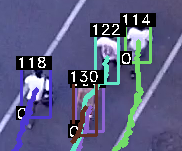} 
  }
  
   \subfigure[] 
   {
  \includegraphics[trim={0.15cm 0 0.25cm 0},clip,height=2cm]{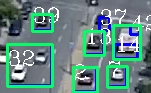} 
  }
   \subfigure[] 
   {
  \includegraphics[height=2cm]{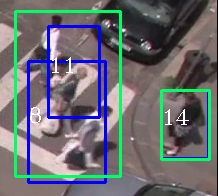} 
}
 \subfigure[] 
   {
  \includegraphics[height=2cm]{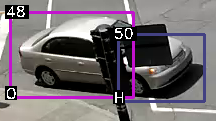} 
 }
  \subfigure[] 
   {
  \includegraphics[height=2cm]{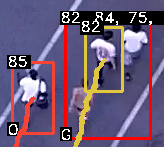} 
}
  \caption{Some examples of our method (first line) and the ViBe algorithm (second line) with both trackers (a), (b), (e) and (f) is MKCF while (c), (d), (g) and (h) is UT. First column is an example of how our method is able to separate close objects in the Rene video. Second column shows how objects moving in opposite directions are easier to separate with our method on the Rouen video. Third column demonstrates how our method is able to keep whole objects in the case of occlusion by other static objects in the Sherbrooke video. Fourth column is an example of how our method gives smaller boxes around the pedestrians in the St-Marc video.}
\label{fig3}
\end{figure*}

\subsection{Edge processing}
\label{edge proc}
This is where we use the background image created at the first step. We extract from the background image $A_i$ the pixels included in the blob $B_i$ followed by an edge detection to form a representation $E_A$ of the same size as $B_i$. We do the same thing for the current image $I$, forming $E_I$ of the same size. Edges of $E_A$ and $E_I$ are obtained using the Canny edge detector\cite{Canny86acomputational} using for threshold the values of $T_H$ and $T_L$ given by
\begin{equation}
T_H = (1 + \sigma) * median(I)
\end{equation}
\begin{equation}
T_L = (1 - \sigma) * median(I)
\end{equation}
The value of $\sigma$ was determined experimentally and set to $\frac{1}{3}$. The value of $median(I)$ is computed for each image $I$ and corresponds to the median pixel value in the grayscale image. 

With the two edges representations for $E_A$ and $E_I$, we can make a logical \textit{xor} operation for each pixel of the regions. This will eliminate the background edges from the foreground blob edges, leaving us with only edges of the foreground $E_F$. Moreover, this enables us to eliminate foreground blobs that were in fact background objects. This increases the precision of our method and adjusts better the size of the detection boxes to the objects. 

Also, the edge processing step can separate two objects or more that are in the same foreground blob $B_i$. This operation will also separate blobs that should not have been merged previously. When we obtain our edge representation image $E_F$, we form groups with the pixels based on distance in order to detect if there are more than one object in the current blob $B_i$. To do this, we choose a random edge pixel $e_i$ and find every connected edge pixel with a Manhattan distance of less than 3 pixels to form an edge group $G_i$. We repeat this process until every edge pixel is a member of an edge group. When this is the case, we find the biggest bounding box $r_i$ for every edge group $G_i$. The number of bounding boxes corresponds to the number of distinct objects in a foreground blob $B_i$. Once again, we store each RoI that has been split into multiple regions in a map for the next step.

\subsection{Decision algorithm}
\label{decision}
At this point, the information we have is three maps: one from the merging of regions by optical flow (see section \ref{merging}), another from the separation of regions (see section \ref{flow sep}) and the last one from the edge analysis (see section \ref{edge proc}). This means that we have, in the best case, two box proposals for each RoI (from the separation map and edge analysis map), but there can be more than that if any of the processing returned more than one box. We present now the algorithm to make our decision regarding which boxes to keep for the final foreground image.

The first thing we check is if an RoI has been modified by the flow separation step (see section \ref{flow sep}). If this is the case, we keep the two boxes returned by the optical flow because we are sure that the two objects were moving in opposite direction. The second verification is if an RoI has been modified both by the flow merging (see section \ref{merging}) and the edges (see section \ref{edge proc}), we keep the boxes from the edges processing, because if the merging was successful, the edges will give us one box around the object and if not, the edges will give us multiple boxes depending on the number of objects. The third check that we make is on the number of boxes that the edge processing step returned. If this number is greater or equal than four, we ignore them and simply keep the one box proposed by the optical flow. This is because it is more likely that the edges over-separated one single object into multiple ones and will lead to a bad detection. The fourth verification covers the situation where the edges proposed two boxes, $e_i$ and $e_k$, and the optical flow only one, $f_k$. This is the hardest case because we do not know if there are truly two objects in the RoI or if, for instance, the edge processing separated the shoes of a pedestrian from the rest of its body. We compute the area ratio, $ratio_{area}$, from both processes in order to make our decision:
\begin{equation}
ratio_{area} = \frac{area(e_i \cup e_j)}{area(f_k)}.
\end{equation}
We keep the the single box $f_k$ from the optical flow processing if the ratio is smaller or equal to 0.65, a parameter determined experimentally. Otherwise, we keep both boxes $e_i$ and $e_j$ from the edges processing step. Finally, in all the other situations, we simply favor the edge boxes over the ones from the optical flow because they tend to be smaller that their counterpart.

\subsection{New final foreground image}
The process to create the new binary image is quite simple. We start by creating an image made only of zero valued pixels. After that, we do a \textit{xor} operation between the image and the box proposals, which are represented by white pixels. This means that when two objects share an intersection, the pixels at the intersection become black and this leads to better detection inputs for the trackers as the objects are separated. The last operation is to increase the size of those intersections by one pixel in every direction (dilation operation) since it facilitates the segmentation for the trackers. Note that since the resulting object masks are combinations of bounding boxes, objects are just segmented coarsely.

\section{Results}
\label{results}
In order to evaluate our proposed method, we used the publicly available UT dataset \cite{DBLP:conf/wacv/JodoinBS14} containing four video sequences of urban mixed traffic. The videos contain pedestrians, cyclists and cars. There were multiple frames that were annotated in each sequence so we could test our method. The evaluation of our method was made in two steps. First, we compared the object detection performance of our method versus the original background subtraction method. Second, we showed how our method can improve the MKCF tracker \cite{DBLP:journals/corr/YangB16a}, a tracker with a simple data association scheme, and the Urban Tracker (UT) \cite{DBLP:conf/wacv/JodoinBS14}, a tracker with a more complex data association scheme, when given the new foreground images compared to the ones produced by the ViBe method~\cite{DBLP:journals/tip/BarnichD11}. Our method improves object detection in all videos, and tracking results for most videos.

The code for our method can be downloaded from \url{https://github.com/beaupreda}.

\subsection{Evaluation methodology}
For the evaluation of our method for the object detection task, we used the Intersection over Union (IoU) metric between the detected bounding boxes and the ground-truth bounding boxes. Then to evaluate our method for the tracking task, we used the tools provided with the Urban Tracker dataset \cite{DBLP:conf/wacv/JodoinBS14}. These tools compute the CLEAR MOT \cite{Bernardin2008} metrics. The multi-object tracking accuracy (MOTA) takes into account the false positives, the ID changes and the misses. The multi-object tracking precision (MOTP) measures the average precision of object matches at each instant. We evaluated our method with an IoU of 30~\%. We decided not to use the classical IoU of 50~\% because when evaluating with the trackers, most of the CLEAR MOT metrics were negatives as the videos are difficult. Also, when looking at the MKCF and UT papers, we found that they were using distances between the centroid of the boxes, and that the values of these distances were quite permissive. For instance, in the Rouen video, the distance threshold was of 164~px, which is 20.5~\% of the width and 27.3~\% of height of the video frame. This distance is generous in a way that objects moderately far away can still be considered matched and tracked. Also, the absolute distance does not consider the size of the objects. For example, there are cars and pedestrians in the Rouen video, so a distance of 164~px might be reasonable for cars that are bigger than pedestrians generally, but not for pedestrians. By using an IoU of 30~\%, we remain flexible for the tracking accuracy while considering the relative size of the different objects that are tracked. We ran the code of both trackers to obtain the results since we changed the evaluation metric. Results are thus different from the ones reported in their respective papers. We kept the default parameters for UT, but had to change the minimum blob size for two videos (Rene-Levesque and St-Marc) for the MKCF tracker. 

For the detection, we also used an IoU of 30~\% because we wanted to remain consistent between our two evaluations. Even when we tested with an IoU of 50~\%, our method had better recall and precision than the original background subtraction.

\subsection{Experimental results}
For the detection task, the results can be found in table~\ref{tab2}. Our method shows improved results for both the precision and recall across all four videos of the Urban Tracker dataset. The most significant improvement is for the Rouen video, followed by Sherbrooke. This can be explained by the fact that a lot of objects are traveling in opposite directions in both of these videos. We are thus able to better separate objects.

\begin{table*}
	\caption{Object detection results of ViBe and our proposed method on the UT dataset. Precision and recall should be high. \textbf{Boldface}: best results}
    \label{tab2} 
	\centering 
	\begin{tabular}{|c|c|c|c|c|}
	\hline
    Dataset & Recall (ViBe) & Recall (Ours) & Precision (ViBe) & Precision (Ours) \\ \hline \hline
    Sherbrooke & 0.606 & \textbf{0.752} & 0.681 & \textbf{0.739} \\ \hline
    Rene-Levesque & 0.812 & \textbf{0.855} & 0.612 & \textbf{0.654} \\ \hline
    Rouen & 0.734 & \textbf{0.834} & 0.724 & \textbf{0.823} \\ \hline
    St-Marc & 0.684 & \textbf{0.754} & 0.415 & \textbf{0.458} \\ \hline
	\end{tabular} 
\end{table*}

The quantitative results for the MKCF tracker and UT are presented in table \ref{tab1}. For the Sherbrooke video sequence, we see that our method is able to improve both the MOTA and the MOTP for both trackers. The MOTA is increased significantly while the MOTP has a more modest improvement. This is due to the fact that the difficulty of this video sequence comes from the large number of cars moving in opposite directions. Our method is able to separate those objects with the optical flow and give an image segmented with each car individually while the original background subtraction merges cars going in opposite direction in the same blob. 

\begin{table*}
   \caption{Tracking results of MKCF and UT using ViBe and our detections on the UT dataset. MOTA and MOTP should be high. \textbf{Boldface}: best results}\label{tab1}
    \centering
    \begin{tabular}{|c|c|c|c|c|c|c|c|c|}
    \hline
    & \multicolumn{4}{c|}{MKCF} & \multicolumn{4}{c|}{UT} \\ \hline
    Dataset & \multicolumn{2}{c|}{MOTA} & \multicolumn{2}{c|}{MOTP} & \multicolumn{2}{c|}{MOTA} & \multicolumn{2}{c|}{MOTP} \\ \hline
    & ViBe & Ours & ViBe & Ours & ViBe & Ours & ViBe & Ours \\ \hline
    Sherbrooke & 0.317 & \textbf{0.523} & 0.553 & \textbf{0.576} & 0.404 & \textbf{0.690} & 0.576 & \textbf{0.590} \\ \hline
    Rene-Levesque & 0.334 & \textbf{0.424} & 0.5309 & \textbf{0.660} & 0.565 & \textbf{0.613} & 0.582 & \textbf{0.705} \\ \hline
    Rouen & 0.501 & \textbf{0.629} & 0.582 & \textbf{0.600} & \textbf{0.696} & 0.670 & 0.617 & \textbf{0.620} \\ \hline
    St-Marc & 0.463 & \textbf{0.534} & \textbf{0.652} & 0.651 & 0.638 & \textbf{0.653} & \textbf{0.691} & 0.682 \\ \hline
    \end{tabular}
 
%
%
\end{table*}
The Rene-Levesque video sequence contains a large number of cars and the camera is far from the scene, which means that the objects of interest are all very small. We increase the MOTA of UT by 5~\% and the MOTP by 12~\%. Our method is able to improve UT because it is able to separate adjacent cars and because the edge processing reduces the size of the boxes. This leads to a more precise tracking. We were also able to improve both metrics with the MKCF tracker, but to do this, we had to reduce the minimum blob size parameter (100 pixels) in the tracker algorithm because, as mentioned earlier, our method reduces the size of the boxes and many proposed boxes were smaller than the threshold originally used by the algorithm. Thus, there were no tracker on some objects which led to poor results. Using the new parameters, we were able to significantly improve the MOTA and MOTP. Note that the same parameters were used with ViBe. 

For the Rouen video, we improve the results of the MKCF tracker in terms of both MOTA and MOTP, while we only improve the MOTP for UT. The difficulty of this dataset is coming from the number of pedestrians crossing the street. Once again, these pedestrians are going in opposite direction but there are also some who walk at the same speed and close to one another. These pedestrians are the hardest to detect individually. Our method, which can segment pedestrians going in opposite directions performs well with the simpler MKCF tracker because it helps the tracker during occlusions. UT remains better with the original background subtraction because our method will sometimes merge two pedestrians going in the same direction.

For the St-Marc video, we also had to change the minimum blob size parameter (700 pixels) in the MKCF tracker for the same reasons as stated above. The MOTA was improved while producing a slightly lower MOTP. The same logic can be transfered for UT where the MOTA was slightly improved and the MOTP decreased. The main challenge from this sequence is that there is a group of four pedestrians walking together. Our method was not able to consistently separate those four pedestrians and this is why we are not able to improve the accuracy by a large margin.

\section{Conclusion}
\label{conclusion}
In this paper, we presented a new method capable of creating better foreground images which, in turn, was shown to improve the performance of two trackers (MKCF and UT). We start from a traditional background subtraction method to obtain our RoI and then, with the help of the optical flow and edge preprocessing, we are able to deal with the fragmentation caused by the background subtraction and effectively separate objects that are either too close to one another or objects that are going in opposite directions. This method improves both the recall and the precision in the object detection task when compared to the original foreground image. It also improves the CLEAR MOT metrics for both trackers for most of the tested videos.

\bibliography{ICPRAI2018_refs}
\bibliographystyle{./IEEEtran}

\end{document}